\pdfoutput=1

\documentclass[11pt]{article}

\usepackage[preprint]{acl}

\usepackage{times}
\usepackage{latexsym}

\usepackage[T1]{fontenc}

\usepackage[utf8]{inputenc}

\usepackage{microtype}

\usepackage{inconsolata}

\usepackage[utf8]{inputenc} 
\usepackage[T1]{fontenc}    
\usepackage{hyperref}       
\usepackage{url}            
\usepackage{booktabs}       
\usepackage{amsfonts}       
\usepackage{nicefrac}       
\usepackage{microtype}      
\usepackage{xcolor}         

\usepackage{booktabs}
\usepackage{arydshln}
\usepackage{graphicx}
\usepackage{subfigure} 
\usepackage{enumitem} 
\usepackage{algorithm}
\usepackage{algpseudocode}
\usepackage{amsmath}
\usepackage{xcolor}

\usepackage{tcolorbox}   
\usepackage{verbatim}    
\usepackage{microtype}
\usepackage{cleveref}

\usepackage{amssymb}

\definecolor{navy}{rgb}{0.0, 0.0, 0.5}

\newenvironment{itemize*}%
{\leftmargini=20pt\begin{itemize}%
\setlength{\itemsep}{3pt}%
\setlength{\parskip}{0pt}%
}%
{\end{itemize}} 
\newenvironment{enumerate*}%
{\begin{enumerate}%
\setlength{\itemsep}{0pt}%
\setlength{\parskip}{0pt}}%
{\end{enumerate}}

\usepackage[dvipsnames]{xcolor}

\newcommand{\think}{\textcolor{NavyBlue}{\textbf{$<$think$>$}}}
\newcommand{\ethink}{\textcolor{NavyBlue}{\textbf{$<$/think$>$}}}

\newcommand{\clarify}{\textcolor{Orange}{\textbf{$<$clarify$>$}}}
\newcommand{\eclarify}{\textcolor{Orange}{\textbf{$<$clarify$>$}}}

\title{SpeakRL: Synergizing Reasoning, Speaking, and Acting \\ in Language Models with Reinforcement Learning}

\author{
Emre Can Acikgoz$^{1}$, Jinoh Oh$^{2}$, Jie Hao$^{2}$, Joo Hyuk Jeon$^{2}$, \\
\textbf{Heng Ji$^{2}$, Dilek Hakkani-Tür$^{2}$, Gokhan Tur$^{2}$, Xiang Li$^{2}$, Chengyuan Ma$^{2}$, Xing Fan$^{2}$}\\
$^{1}$University of Illinois Urbana-Champaign, $^{2}$Amazon Alexa\\
\normalsize{\texttt{acikgoz2@illinois.edu},\; \texttt{ojino@amazon.com}}\\
}

\begin{document}
\maketitle
\begin{abstract}

Effective human-agent collaboration is increasingly prevalent in real-world applications. Current trends in such collaborations are predominantly unidirectional, with users providing instructions or posing questions to agents, where agents respond directly without seeking necessary clarifications or confirmations.
However, the evolving capabilities of these agents require more proactive engagement, where agents should dynamically participate in conversations to clarify user intents, resolve ambiguities, and adapt to changing circumstances. Existing prior work under-utilize the conversational capabilities of language models (LMs), thereby optimizing agents as better followers rather than effective speakers.
In this work, we introduce \textbf{SpeakRL}, a reinforcement learning (RL) method that enhances agents' conversational capabilities by rewarding proactive interactions with users, such as asking right clarification questions when necessary.
To support this, we curate \textbf{SpeakER}, a synthetic dataset that includes diverse scenarios from task-oriented dialogues, where tasks are resolved through interactive clarification questions.
We present a systematic analysis of reward design for conversational proactivity and propose a principled reward formulation for teaching agents to balance \textit{asking} with \textit{acting}.
Empirical evaluations demonstrate that our approach achieves a 20.14\% absolute improvement in task completion over base models without increasing conversation turns even surpassing even much larger proprietary models, demonstrating the promise of clarification-centric user-agent interactions.

\end{abstract}

\begin{figure*}[h!]
    \centering
    \includegraphics[width=\textwidth]{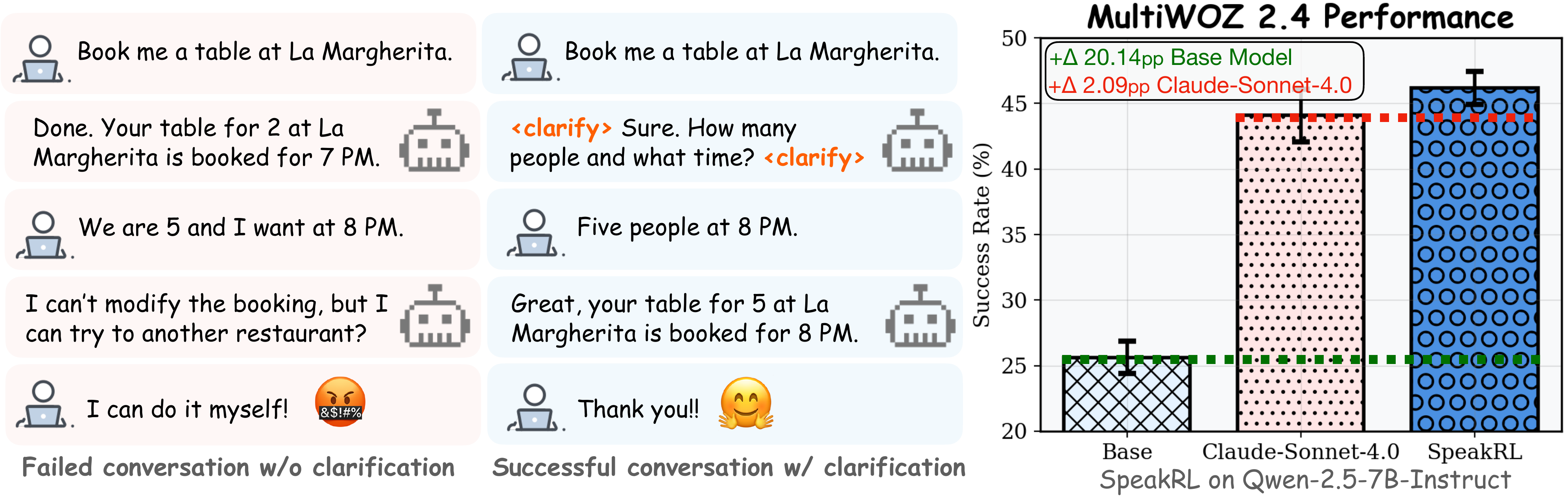}
    \vspace{-7mm}
    \caption{\textbf{Impact of SpeakRL on Success Rate Performance in MultiWOZ 2.4.} \textbf{Left:} Example dialogues showing failure without clarification (left) versus success with proactive clarification (right). \textbf{Right:} Success rates for the Base Model (25.63\%), Claude-Sonnet-4.0 (44.08\%), and SpeakRL (46.17\%) on MultiWOZ 2.4. SpeakRL attains ~80\% higher success rate than the Base Model and $\sim$5\% higher than Claude-Sonnet-4.0, demonstrating the substantial impact of reinforcement-learned clarification in multi-turn conversation settings.}
    \label{fig:fig1}
    \vspace{-5mm}
\end{figure*}

\vspace{-3mm}
\section{Introduction}

The integration of language models (LMs) into real-world applications has transformed human–agent collaboration, enabling systems that assist users with tasks ranging from planning travel itineraries~\cite{xie2024travelplanner, yao2024taubench} to managing smart home ecosystems~\cite{gottardi2022alexa, padmakumar2022teach}. However, during these interactions, agents often encounter vague or underspecified user queries, making task execution more challenging. In such situations, the agent must either make a potentially risky assumption or fail to complete the task~\cite{purver2001clarificationdialogue}. 

Mechanistically, asking clarification questions serves as a proactive error-correction mechanism in conversational agents: by querying for missing details early, agents minimize uncertainty, refine task understanding, and prevent downstream failures~\citep{acikgoz2025desideratum}. This dynamic is illustrated in Figure~\ref{fig:fig1}, where an ambiguous restaurant booking request fails without clarification (left) but succeeds when the agent seeks key details (right), highlighting clarification's role in robust, multi-turn dialogues. Thus, we treat clarification as a control primitive: detect underspecification, ask for the missing variables, then execute. This loop grounds actions in user constraints—reducing plan entropy, avoiding risky commitments, and yielding robust task completion. 

Existing methods for integrating LMs with clarification capabilities generally fall into two categories: (i) designing hand-crafted, rule-based prompts with predefined instructions~\citep{dongre2024respact} and (ii) fine-tuning models explicitly to generate clarification questions~\citep{zhang2025intentsim, zhang2025modelingiclr} for better interactions with users~\citep{li2025gate, andukuri2024stargate}. In parallel to these, reinforcement learning (RL)~\cite{sutton2018rl} has gained significant traction in improving reasoning capabilities in Large Reasoning Models (LRMs) like OpenAI-o1~\cite{jaech2024openai} and DeepSeek-R1~\cite{guo2025deepseekr1}, employing techniques such as GRPO~\cite{shao2024grpo} to enhance problem-solving skills through experiential reward feedback. 

However, applying reinforcement learning with verifiable rewards (RLVR) to interactive user clarification remains underexplored and introduces several challenges: (i) \textbf{Multi-turn Conversation and Clarification}—the agent must balance between directly responding and selectively requesting clarification only when ambiguity arises, which demands sophisticated multi-turn interaction capabilities; (ii) \textbf{Reward Design}—creating an effective reward function that clearly defines when and what to ask for clarification remains challenging, as it is uncertain whether simple outcome-based rewards can guide agents to consistently and meaningfully generate clarification questions; and (iii) \textbf{RL Optimization}—integrating clarification behaviors into RL training for LLMs in a stable and efficient manner is still an open problem.

To address the aforementioned challenges, we introduce SpeakRL, a novel RLVR algorithm that empowers LLMs to resolve ambiguity through user-directed clarification in multi-turn conversations by learning both \textit{when} and what to \textit{ask} to fill particular slots~\citep{lison2013model, louvan2020slotfilling}. Our SpeakRL uses Group Relative Policy Optimization (GRPO), an on-policy RL algorithm, with two complementary verifiable reward signals. We first introduce structured special tokens that separates internal uncertainty reasoning (\think…\ethink) and clarification questions (\clarify…\eclarify), giving the agent precise control over the timing of clarification requests by learning to produce these tokens appropriately. Second, we define an LLM-as-judge reward model that evaluates and optimizes the quality of clarification questions, teaching the model to formulate more effective queries. Together, this RLVR-based optimization enables agents to learn strategic question-asking behavior without explicit task completion rewards. Notably, by optimizing solely for clarification quality rather than task success, we demonstrate that effective interactive conversation naturally leads to more successful and efficient task completion.
 

In summary, our primary contributions are as follows: (1) We introduce SpeakRL, an end-to-end RLVR framework that enables LLM agents to iteratively improve their ability to ask clarification questions in multi-turn, goal-oriented dialogues. (2) We construct SpeakER, a synthetic dataset of 25,000 task-oriented multi-turn conversations, explicitly designed to include ambiguous scenarios annotated with user clarification turns. (3) We design reward strategies within RLVR that guide agents on both when and what clarification questions to ask. (4) We show that post-training with SpeakRL enables LLMs to proactively ask clarification questions in uncertain or ambiguous contexts, improving task success while reducing dialogue length, thereby fostering more accurate and efficient collaborative human–agent interactions.

\section{Related Work}

\noindent\textbf{Reinforcement Learning for Task-Oriented Dialogue.\;\;} RL has been applied to learn dialogue behaviors beyond supervised imitation, from optimizing open-ended generation with long-horizon rewards \citep{li-etal-2016-deep} to traditional TOD agents that act for information access \citep{dhingra-etal-2017-towards}. Prior work also explores interactive improvements via self-play and online RL~\citep{shah-etal-2018-bootstrapping}, collaborative multi-agent RL dialogue training~\citep{papangelis-etal-2019-collaborative}, and the importance of action-space design for effective RL~\citep{zhao-etal-2019-rethinking}. Motivated by the fact that modern LLM-based TOD systems are end-to-end~\citep{11165061}, we use RLVR to directly reward clarification behavior in natural language, enabling a simple and end-to-end pipeline that teaches the model when and how to clarify without hand-crafted states or rigid dialog acts.

\noindent\textbf{Reinforcement Learning and LLMs.\;\;}
RL has been incorporated into LLMs mainly via Reinforcement Learning from Human Feedback (RLHF), which trains a reward model from human preferences and optimizes the policy using PPO \citep{ouyang2022rlhf, schulman2017ppo}. However, PPO is often unstable and requires careful hyperparameter tuning. To mitigate these issues, simpler alternatives such as Direct Preference Optimization (DPO) have been proposed, which learn directly from preference pairs without explicit reward modeling, along with several efficient variants \citep{rafailov2023dpo, ethayarajh2024kto, hong-etal-2024-orpo, meng2024_simpo}. Although these methods improve computational efficiency, they are prone to off-policy issues and often fall short of the performance achieved by traditional RL techniques~\citep{pang2024iterativeprefence}. More recently, Group Relative Policy Optimization (GRPO) has been proposed~\citep{shao2024deepseekmath, guo2025deepseekr1}, which bypasses the need for a reward model by employing a group-based evaluation approach and demonstrates robust enhancements in reasoning capabilities across diverse tasks and domains~\citep{qian2025toolrl, jin2025searchr1, lai2025medr1, huang2025visionr1}. Nevertheless, the use of RL to train conversational agents for greater proactivity remains a largely untapped area of research.

\vspace{1mm}

\noindent\textbf{Asking Clarification Questions.\;\;}
Prior work addresses ambiguity in user requests by teaching LLMs to ask clarification questions, using either prompting-based approaches with hand-engineered instructions~\citep{zhang2025intentsim, dongre2024respact} or explicit training methods~\citep{zhang2025modelingiclr, andukuri2024stargate, wu2025collabllm, chen2025learning, kobalczyk2025active}. Training-based methods employ various techniques, including supervised fine-tuning (SFT)~\citep{andukuri2024stargate}, reinforcement learning~\citep{chen2025learning, wu2025collabllm}, direct preference optimization with positive and negative samples~\citep{zhang2025modelingiclr, chen2025learning}, and active learning~\citep{kobalczyk2025active}.
However, most of these approaches focus primarily on clarification question generation in isolation and underutilize the complexity of multi-turn conversational dynamics, with notable exceptions being~\citet{dongre2024respact} and~\citet{wu2025collabllm}. \citet{dongre2024respact} explores multi-turn settings but relies on hand-engineered prompts where speaking actions are conditioned as policies for specific situations, limiting generalizability. Closest to our work, \citet{wu2025collabllm} train LLMs to ask clarification questions in multi-turn settings using user simulators and reward signals. However, their setup is domain-general and does not require task completion or agentic behaviors such as tool use, limiting realism for task-oriented settings.
In contrast, we apply RLVR in multi-turn TOD with user feedback tied to task completion, requiring function calling and tool use to achieve realistic agentic behavior.

\section{SpeakRL}
\label{sec:speakrl}

\begin{figure*}[h!]
    \centering
    \includegraphics[width=\textwidth]{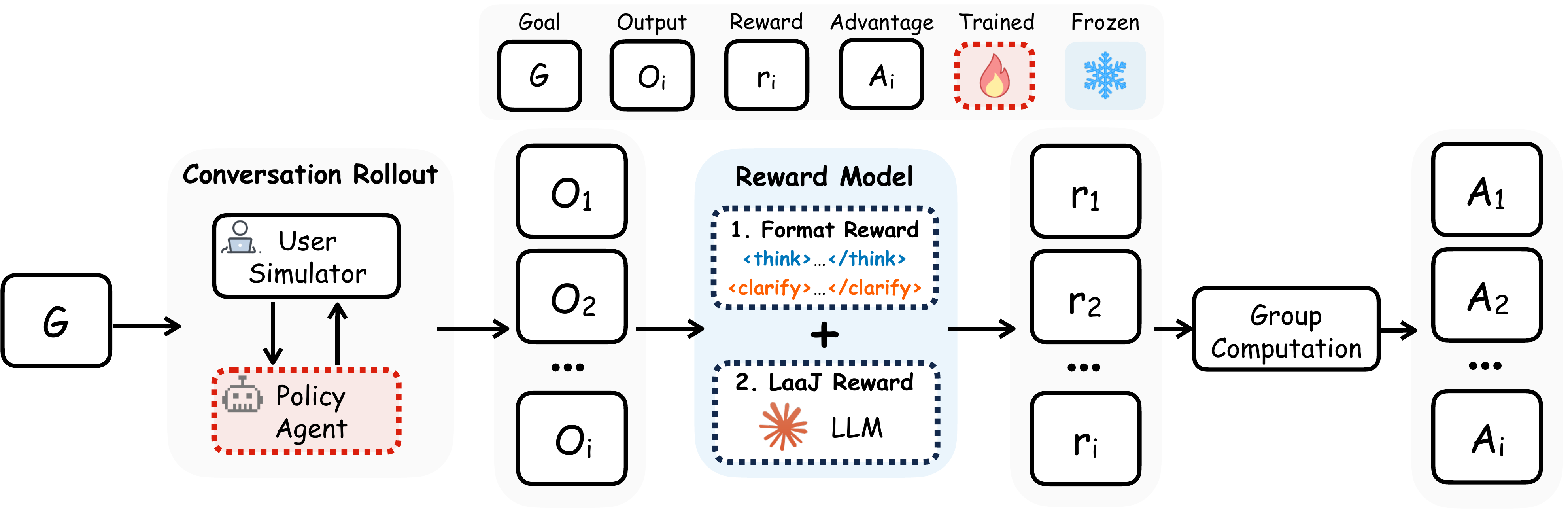}
    \vspace{-7mm}
    \caption{\textbf{GRPO algorithm with user clarification for collaborative RL.} 
    Demonstration of GRPO training for teaching asking clarification questions (SpeakRL). During rollout, the policy agent conducts multi-turn interactions, with outputs rewarded according to format compliance and LLM-as-a-Judge (LaaJ) scores.}
    \vspace{-5mm}
    \label{fig:grpo}
\end{figure*}

\noindent\textbf{Overview.} On high level, SpeakRL teach LLM Agents to identify when user requests are underspecified or ambiguous and to ask targeted clarification questions that maximize task success in goal-oriented dialogues.
Each conversation begins with a user request that the agent attempts to fulfill through iterative interaction, potentially spanning multiple related tasks (e.g., purchasing train tickets followed by booking nearby accommodations). 
For actual realistic conversations as in real-life, we simulate human users through a user simulator that takes both a goal message and a role-defining prompt as input~\citep{xu2024autotod}.
The agent, instantiated as an LLM, iteratively reasons about whether a clarification question is necessary and, if so, formulates the appropriate query. 
Over time, the agent progressively learns to enhance its internal reasoning abilities~\citep{zelikman2022star}, allowing it to better grasp ambiguities and determine what to ask, thereby improving the quality of its clarification questions.
Through this iterative process, the agent improves both its reasoning capabilities and clarification question quality. 
Conversations terminate either when the agent completes the task and signals completion, or when the maximum dialogue length of 20 turns is reached.

\subsection{Task Definition}
Task-oriented dialogues (TOD) can be viewed as \textit{multi-step reasoning processes} where an agent interacts with a user (or user simulator) to accomplish a concrete goal—such as booking, ordering, or scheduling—through successive decisions and clarifications. Each decision in the dialogue depends on the accumulated interaction history, and the final outcome is determined by the cumulative success of these intermediate reasoning steps.

Formally, let the agent’s action space be:
\begin{equation}
    \mathcal{A} = {a_1, a_2, \dots, a_n},
\end{equation}
where each action $a_i \in \mathcal{A}$ corresponds to one of three categories: (i) asking a clarification question, (ii) generating an actual response, or (iii) executing an API call. Given a user goal $\mathcal{G}$, the dialogue trajectory up to step $k$ is defined as:
\begin{equation}
    s_k = (r_1, a_1, o_1), \dots, (r_k, a_k, o_k),
\end{equation}
where $r_i$ represents the agent’s reasoning or internal plan at step (i), $a_i$ denotes the chosen action, and $o_i$ represents the observation received after executing $a_i$, which may include user or environment feedback.

At each step $k+1$, the agent interprets the current dialogue state, generates the next reasoning step $r_{k+1}$, selects an action $a_{k+1} \in \mathcal{A}$, and produces the corresponding utterance or API call to advance toward fulfilling $\mathcal{G}$. The agent’s policy is defined as:
\begin{equation}
    \pi : s_k \rightarrow (r_{k+1}, a_{k+1}),
\end{equation}
with the objective of selecting the optimal action that maximizes expected reward:
\begin{equation}
    a^*_{k+1} = \arg\max_{a_{k+1} \in A} R(s_k, a_{k+1}, o_{k+1}),
\end{equation}
where $\mathcal{R}(\cdot)$ evaluates progress made after performing the action—reflecting factors such as effective clarification, correct slot acquisition, or successful task advancement.

While immediate rewards encourage effective reasoning and interaction at each step, the policy $\pi$ is optimized to maximize the cumulative reward across the dialogue trajectory:
\begin{equation}
    \max_{\pi} \; \mathbb{E}_{\pi} \sum_{k=1}^{K} R(s_k, a_{k+1}, o_{k+1}),
\end{equation}
This step-wise optimization enables the agent to learn both \textit{when to ask } and \textit{what to ask}, balancing clarification with progression toward the final goal. Through reinforcement signals, the agent learns to navigate the trade-off between proactive information gathering and efficient task completion, ultimately leading to more robust and goal-aligned dialogue behavior.

Importantly, we focus on clarification triggered by referential ambiguity or underspecification, where multiple valid slot values remain plausible despite partial information rather than traditional slot filling for missing required fields~\citep{lison2013model, louvan2020slotfilling}, and train agents to decide when uncertainty warrants clarification rather than simply requesting unfilled slots.

\subsection{Structured Reasoning and Clarification Tokens}
To enable the model to autonomously reason about ambiguity and generate clarification questions, we structure its outputs using two category of special tokens: \think…\ethink\ and \clarify…\eclarify. The \think\ tokens delimit the model’s internal reasoning trace, allowing it to articulate latent uncertainty and evaluate whether the current user input provides sufficient information to act. 
\clarify\ tokens, in turn, marks the model’s externally verbalized clarification question aimed at resolving that uncertainty. 
This tokenization provides a clean separation between implicit reasoning and explicit interaction, enabling precise supervision and reward assignment during RL training.

When the model emits a segment within \think…\ethink\, the content is treated as an internal thought process and excluded from the dialogue context visible to the user. 
If the output contains \clarify…\eclarify, the enclosed text is parsed as the model’s clarification question and appended to the dialogue history, triggering a response from the user simulator. 
The returned feedback is then incorporated into the evolving dialogue state, forming a new step in the reasoning trajectory.

Importantly, \think\ and \clarify\ can co-occur within a single output, allowing the model to reason, identify uncertainty, and immediately issue a targeted clarification within the same turn. The user’s initial goal or query $\mathcal{Q}$ is provided as the starting context, and subsequent user replies are iteratively appended to form a structured multi-turn trajectory:
\begin{equation}
    s_k = (r_1, a_1, o_1), \dots, (r_k, a_k, o_k),
\end{equation}
where reasoning $r_i$ corresponds to \think\ content, and clarification or response $a_i$ corresponds to user-directed actions (via \clarify\ or plain responses).

This token-level design enables reinforcement signals to be applied at fine granularity, rewarding the model not merely for end-task success but for strategic \textit{decision-making} in ambiguity detection and question formulation. Through this structured reasoning–clarification loop, SpeakRL teaches LLMs to proactively manage uncertainty and conduct effective multi-turn dialogue grounded in user intent.

\subsection{Reward Design}
Reward mechanisms play a central role in reinforcement learning with verifiable rewards (RLVR), guiding the model toward desirable interactive behavior~\citep{guo2025deepseekr1}. In our training, we similarly adopt a reward formulation that integrates structural and semantic-quality components~\citep{jin2025searchr1, qian2025toolrl}, implicitly teaching the model when to ask for clarification via token-level optimization, and what to ask through semantic feedback. Formally, the total reward at each step is defined as:
\begin{equation}
    R_{\text{total}} = R_{\text{format}} + R_{\text{clarify}},
\end{equation}
where ($R_{\text{format}}$) measures adherence to the required output structure and ($R_{\text{clarify}}$) assesses the quality and helpfulness of clarification questions.

\begin{table*}[t]
\centering
\begin{tabular}{p{0.9\linewidth}}
\toprule
Given the \textbf{Conversation} below, carefully read the dialogue and the final user query. First, reflect on the reasoning process—consider any ambiguity, missing information, or potential failure points. Then decide whether it is necessary to ask the user a clarification question before proceeding.
The reasoning process and user clarification question are enclosed within \think…\ethink\ and \clarify…\eclarify\ tags, respectively, i.e., \think reasoning process here \ethink\ \clarify\ user clarification question here \eclarify. User: \textcolor{red}{conversation}. Agent: \\
\bottomrule
\end{tabular}
\caption{\textbf{Template for SpeakRL.} The placeholder \textcolor{red}{conversation} is substituted with the corresponding user query and dialogue turn during training.}
\label{tab:deepseek_template}
\end{table*}

\begin{figure*}[h!]
    \centering
    \includegraphics[width=\textwidth]{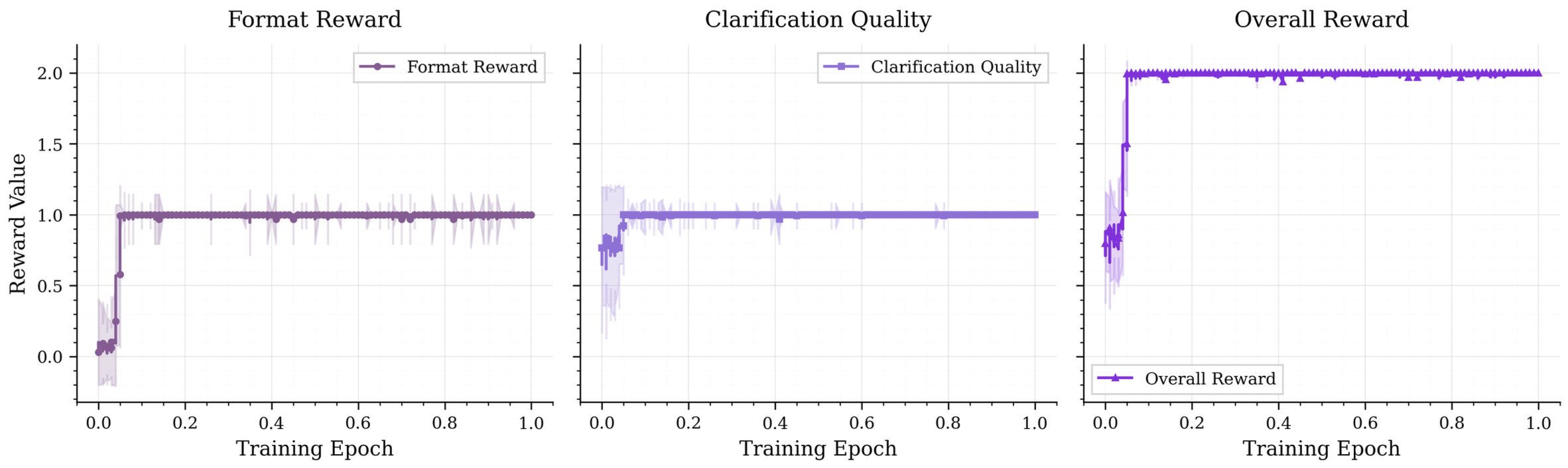}
    \vspace{-5mm}
    \caption{\textbf{RL Rewards}. SpeakRL reward progression across format, clarification quality, and overall.}
    \vspace{-3mm}
    \label{fig:fig4}
\end{figure*}

\vspace{3mm}

\noindent\textbf{Format Reward.\;\;} The format reward ($R_{\text{format}}$) verifies whether the model correctly employs the designated special tokens \think\ and \clarify\ in the proper order and syntactic form. The reward encourages the model to produce interpretable reasoning traces and explicitly structured clarifications, where a fixed output format simplifies verification and guides the model toward more deliberate reasoning.
\begin{equation}
    R_{\text{format}} =
\begin{cases}
1, & \text{\small if tokens appear correctly in valid order,}\\
0, & \text{\small otherwise.}
\end{cases}
\end{equation}
Additionally, partial credit can be assigned when the model produces only one of the required fields (e.g., emits \think\ but omits \clarify), which empirically stabilizes early-stage learning. This binary (or near-binary) format supervision ensures the model first learns how to produce syntactically valid clarification outputs before optimizing their content quality.

\vspace{3mm}

\noindent\textbf{Clarification Reward.\;\;} The clarification reward $R_{\text{clarify}}$ evaluates the \textit{semantic quality} of the clarification question enclosed within \clarify…\eclarify. Because there is no single “correct” clarification for a given ambiguous query, we adopt an LLM-as-judge scoring framework that assigns verifiable feedback based on several interpretable dimensions. At each clarification step, we query an LLM with a structured rubric prompt that evaluates complementary dimensions of clarification quality as defined in \Cref{prompt:llm_judge_prompt}. Specifically, Relevance measures whether the question directly targets the ambiguous or missing information in the user request; Precision and Clarity capture whether the question is well-formulated, unambiguous, and clearly phrased; Specificity assesses whether it narrows down the uncertainty to a concrete aspect of the task; Logical Connection evaluates whether the question follows coherently from the preceding dialogue context; and Constructive Nature examines whether the question helps advance task completion rather than repeating or restating information.

The combined reward is thus expressed as: $R_{\text{total}} = R_{\text{format}} + R_{\text{clarify}}$. Unlike prior works that rely solely on outcome-based or rule-matching rewards, our design jointly optimizes structural adherence and interaction quality. The format reward ensures syntactic precision in the model’s reasoning–clarification structure, while the clarification reward provides dense, interpretable feedback on what the agent asks. This two-part signal allows SpeakRL to learn fine-grained clarification behaviors without relying on ground-truth task completion labels, ultimately leading to more adaptive, user-aware dialogue strategies.

\subsection{SpeakER Dataset}

We introduce the SpeakER dataset to study clarification behavior in task-oriented dialogue settings where user goals are \emph{intrinsically ambiguous} and cannot be resolved without explicitly asking clarification questions. Unlike conventional slot-filling datasets (e.g., SLURP), where missing information can be obtained by sequentially filling required slots, SpeakER focuses on scenarios where the system has \emph{partial but uncertain} information, and successful task completion depends on resolving ambiguity through clarification.

We build SpeakER by using MultiWOZ 2.4~\citep{ye2022multiwoz24} dialogues as seed trajectories and synthesizing new dialogue paths that intentionally introduce ambiguities at different stages of the interaction. These ambiguities may require one or multiple clarification turns to resolve. Clarification turns are explicitly annotated using special tokens \clarify...\eclarify, enabling turn-level supervision of \emph{when} clarification is necessary within a multi-turn context. This annotation allows us to condition learning on the full dialogue history while assigning rewards at specific clarification decision points, supporting single-step optimization with multi-turn conversational context. Importantly, SpeakER is not designed to optimize \emph{what} clarification question to ask, but rather to supervise \emph{whether} and \emph{when} clarification is required; the former is studied separately in \Cref{sec:speakrl}. All dialogues are filtered to remove redundant clarification questions using n-gram similarity, and only trajectories that successfully complete the task through clarification are retained. The final dataset consists of approximately 25K training dialogues.

For preference-based training, we additionally construct SpeakER-DPO. Positive samples correspond to successful clarification-based trajectories, while negative samples reuse the same user goals but omit clarification turns, leading to task failure. We use DPO~\citep{rafailov2023dpo} as a contrastive objective rather than human preference alignment, allowing the model to learn the consequences of asking—or failing to ask—clarification questions. All data are synthesized using \texttt{claude-sonnet-4-20250514}.

\subsection{RL Training}
To train the model with structured rewards, we adopt the GRPO algorithm~\citep{shao2024deepseekmath} by using the training instruction in \Cref{tab:deepseek_template} (See \Cref{fig:grpo}). Unlike the original formulation, we remove the KL-divergence penalty against a reference model, allowing the policy to more freely adapt to our custom clarification format and reward structure. This design choice simplifies the training pipeline while maintaining stability and leading to faster convergence in practice. For the clarification-quality reward, we experiment with two LLM-as-judge settings: a strong external evaluator, \texttt{claude-sonnet-4-20250514}, providing high-fidelity feedback for objective scoring; and a self-judging setup, where the same model \texttt{Qwen2.5-7B-Instruct} evaluates its own clarification questions. The latter explores the potential of self-improving agents that refine their behavior through internally generated reward signals~\citep{huang2023selfimprovellm, huang2025selfimprovementsharp, acikgoz2025ttsi}.

During RL training, the model rapidly learns both structured and behavioral reward signals on SpeakER. As shown in \Cref{fig:fig4}, the \textit{Format Reward} starts low but quickly converges to a stable maximum, indicating that the model efficiently learns to follow the expected \think\ and \clarify\ output structure. The \textit{Clarification Reward} (middle) begins at a moderately higher baseline and similarly converges early, suggesting that the agent quickly internalizes what constitutes an effective clarification. Together, these trends yield a stable \textit{Overall Reward} (right), demonstrating consistent convergence and stable policy improvement throughout GRPO training.

\section{Main Results}

\noindent\noindent\textbf{Collaborative Environment.\;\;} 
To simulate realistic user environments, agents must engage in collaborative communication that handles real-world goal-oriented tasks. We simulate conversations between an agent and a human user-simulator~\citep{xu2024autotod} with access to user goals hidden from the agent. The agent must fulfill user requests that may span multiple subtasks (e.g., booking a hotel, finding an Italian restaurant, and reserving a table for 3 at 7pm), some containing ambiguities requiring clarification. The agent must interact with the user to gather necessary information and complete the task. Task completion occurs when the agent returns correct booking or reservation IDs, or terminates after a predefined turn limit. We evaluate the agent’s performance using two key metrics: success rate and average number of turns where a lower value indicates better efficiency (See \Cref{app:env} for further details).

\vspace{2mm}

\noindent\textbf{Models.\;\;} We use \texttt{Qwen2.5-7B-Instruct} as our main agent model because it is publicly available as open source, has been shown to be one of the best models for its size, and is generally used in RL fine-tuning. We use \texttt{claude-sonnet-4-20250514} for the user-simulator and LaaJ reward model unless otherwise specified.

\begin{table}[t]
\centering
\resizebox{\columnwidth}{!}{%
\begin{tabular}{lcc}
\toprule
\textbf{Method}        & \textbf{Success ($\uparrow$)}    & \textbf{Turns ($\downarrow$)} \\ \midrule
\textit{Qwen-2.5-7B-Instruct}                                      &                          &                 \\
\hspace{4mm}Prompting                               & 25.63 \small ±  1.24                         & 8.12                \\
\hspace{4mm}SFT                                     & 28.78 \small ± 1.15                         & 7.32                \\
\hspace{4mm}DPO                                     & 45.73 \small ± 3.23                           & 5.92                \\
\hspace{4mm}SpeakRL                                       & \textbf{46.17 \small ± 1.25}               & \textbf{5.82}               \\ \hdashline[0.5pt/2pt] \addlinespace[1mm]

claude-sonnet-4.0                                 & 44.08 \small ± 1.99            & 6.28 \\ \bottomrule
\end{tabular}%
}
\caption{\textbf{Main results on collaborative user–agent dialogue evaluation.} Comparison of different training paradigms on task success (Success) and conversational efficiency (\textit{Avg. Turns}). Higher success and fewer turns indicate better goal completion and dialogue quality.}
\vspace{-5mm}
\label{tab:results}
\end{table}

\vspace{2mm}

\noindent\textbf{Finding 1: Effective User Clarification Improves Task Success and Efficiency.\;\;}
Even though SpeakRL is not explicitly optimized for task success, it achieves substantial improvements over prompting. Specifically, Success (Avg@5) improves from 25.63 $\rightarrow$ 46.17, corresponding to an absolute gain of +20.54 points (80\% relative). These gains indicate that effective clarification, asking when information is missing, directly enhances task completion rates. Moreover, SpeakRL reduces average turns from 8.12 $\rightarrow$ 5.82, a reduction of 2.30 turns (28\%), demonstrating improved conversational efficiency. The agent learns to identify ambiguities early, ask a single targeted question, and obtain the necessary information in fewer exchanges, minimizing unnecessary dialogue cycles.

\vspace{2mm}

\noindent\textbf{Finding 2: GRPO-Based Reinforcement Learning Outperforms Supervised and Preference-Based Methods.\;\;}
Among learning paradigms, GRPO-based SpeakRL achieves the strongest performance, outperforming both SFT and DPO. Compared to SFT, Success rises from 28.78 $\rightarrow$ 46.17, an absolute +17.39 gain (relative 60\%). SFT overfits to dialogue trajectories--imitating structure without learning when or why to ask questions—whereas SpeakRL learns through reward feedback. Against DPO, SpeakRL still achieves higher scores (45.73 $\rightarrow$ 46.17), showing the benefit of granular, token-level reward shaping via GRPO. These results highlight that structured RL-based reward learning can produce reasoning-capable and adaptive conversational agents beyond imitation or pairwise preference optimization.

\vspace{2mm}

\begin{figure*}[t!]
    \centering
    \includegraphics[width=\textwidth]{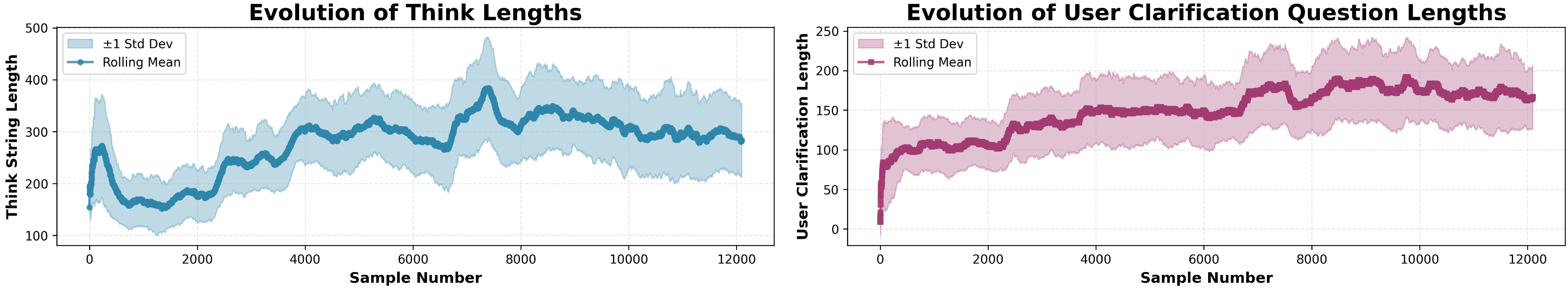}
    \caption{\textbf{Evolution of reasoning and clarification behaviors during GRPO training.} The left plot tracks the growth of \think\ sequence lengths, reflecting deeper internal reasoning, while the right plot shows increasingly rich \clarify\ questions, indicating the model’s improved ability to identify and resolve missing information.}
    \label{fig:fig2}
\end{figure*}

\begin{figure*}[t!]
    \centering
    \includegraphics[width=\textwidth]{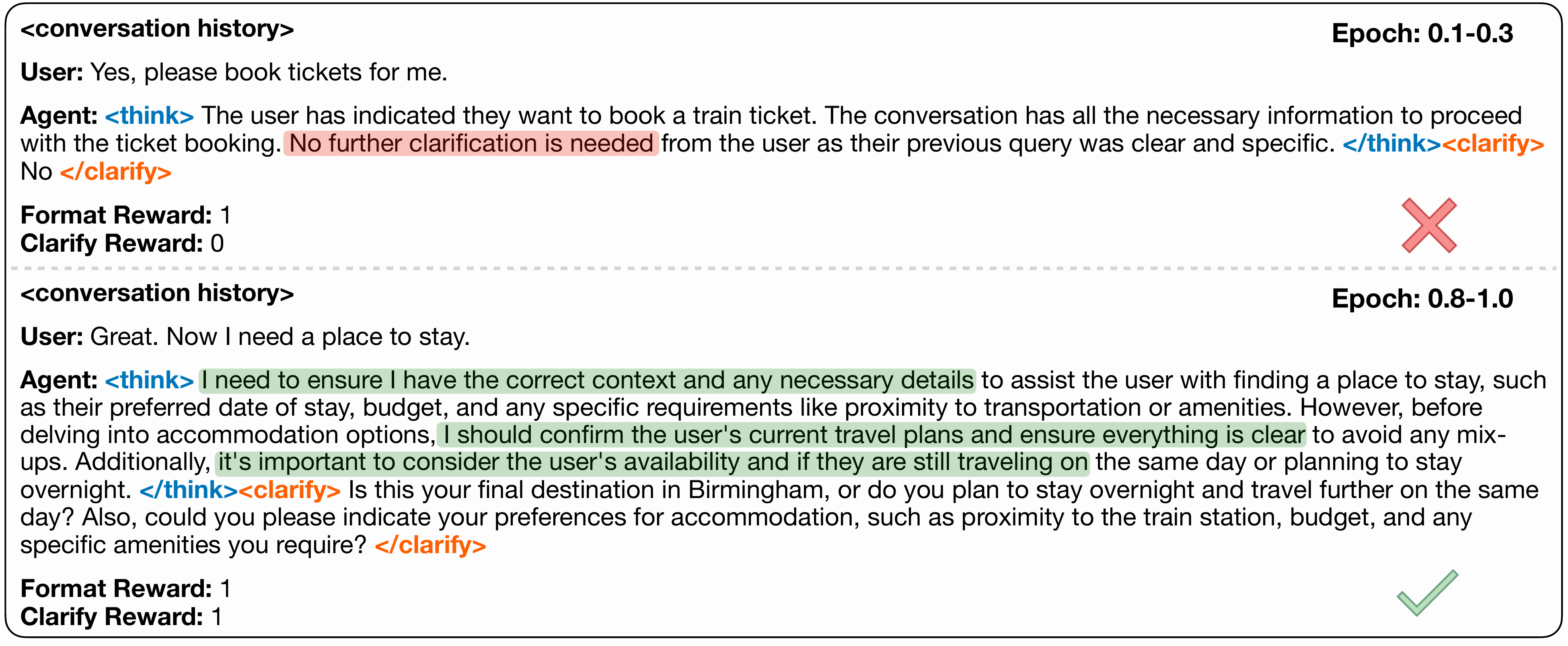}
    \vspace{-5mm}
    \caption{\textbf{Qualitative Analysis of Thinking and Clarification.} Early in training (Epoch 0.1–0.3), the agent prematurely concludes that no clarification is needed, overlooking missing context. Later (Epoch 0.8–1.0), it demonstrates careful thinking, confirming user goals, and generating clear targeted clarification questions.}
    \vspace{-3mm}
    \label{fig:fig3}
\end{figure*}

\noindent\textbf{Finding 3: Small Open Models Can Surpass Proprietary LLMs When Trained Collaboratively.\;\;}
Remarkably, SpeakRL fine-tuned on the open-source \texttt{Qwen2.5-7B-Instruct} (46.17\%) surpasses much more larger proprietary model \texttt{claude-sonnet-4.0} (44.08\%). Despite being significantly smaller, SpeakRL benefits from reinforcement-driven clarification training, enabling it to generalize beyond memorization and achieve competitive or superior task success. This finding underscores the promise of small, open, and interpretable conversational agents~\citep{belcak2025small}, when trained with collaborative user-clarification feedback through RLVR, to rival and even outperform larger closed models.

\section{Ablation Studies}

We conduct detailed analyses to understand the internal components of SpeakRL and the underlying dynamics of the RLVR process.

\vspace{2mm}

\noindent\textbf{Finding 4: Emergent Improvement in Latent Reasoning Depth During RLVR Training.\;\;}
As shown in \Cref{fig:fig2} (left), although the \think...\ethink\ token sequence is not explicitly rewarded for its length, we observe a gradual and consistent increase in the model’s internal reasoning span over RLVR iterations. The average think-string length steadily rises throughout training, indicating that the model autonomously learns to engage in deeper reasoning before producing actions or clarifications. This emergent behavior reveals that GRPO not only optimizes for external task success but also implicitly fosters the development of richer internal deliberation, leading to improved reasoning quality and more stable decision-making over time.

\vspace{2mm}

\noindent\textbf{Finding 5: Learning What to Ask Leads to Richer and More Effective Clarifications.\;\;}
As shown in \Cref{fig:fig2} (right), the model gradually learns \textit{what to ask}, how to identify and query missing information critical for task completion. Early in training (first 100–200 samples), clarification questions are short and underspecified (around 10 tokens), often failing to resolve ambiguity. Over time, their average length increases steadily, indicating that the agent begins forming more complete and contextually grounded questions. This evolution demonstrates that reinforcement learning not only improves the model’s ability to act but also shapes its inquiry behavior, enabling it to formulate richer, more purposeful clarifications that directly enhance task success and collaborative efficiency.

\subsection{Qualitative Analysis}

As shown in \Cref{fig:fig3}, our qualitative analysis uncovers an emergent pattern of reflective reasoning within the \think...\ethink\ segments, revealing how the model progressively internalizes the principles of context-aware clarification through RL. Early in training (Epoch 0.1–0.3), the agent’s reasoning remains superficial; its thought process ends prematurely with conclusions such as “The conversation has all the necessary information… No further clarification is needed”.  It incorrectly concludes that no clarification is needed, even though additional information is clearly required. By contrast, in later stages (Epoch 0.8–1.0), the model’s internal reasoning exhibits a more structured and anticipatory nature. It begins to self-monitor and generate meta-cognitive statements such as “I need to ensure I have the correct context and any necessary details” and “I should confirm the user’s current travel plans and ensure everything is clear to avoid any mix-ups”. These phrases indicate that the model is learning to (1) assess the sufficiency of information, (2) reason about latent variables like time, intent, and user preferences, and (3) plan clarification queries that minimize ambiguity before acting. 

\section{Discussion}

\paragraph{Conclusions}
In this work, we presented SpeakRL, an end-to-end RLVR framework that enables LLM agents to proactively ask effective clarification questions in multi-turn, goal-oriented dialogues. To do that we create SpeakER, a synthetic dataset of 25K conversations explicitly designed to capture ambiguous scenarios through turn-level clarification annotations. By separating reasoning and clarification using structured tokens and train with GRPO-based RLVR, SpeakRL jointly learns when and what to clarify without directly optimizing for task completion. Empirical results demonstrate that post-training with SpeakRL leads to higher task success and shorter dialogues, resulting in more accurate, efficient, and collaborative human–agent interactions.

\paragraph{Limitations}
While SpeakRL shows promise for co-evolving user–agent interactions, it has several limitations. First, both training and evaluation rely on the training split of MultiWOZ 2.4 due to the lack of suitable task-oriented user simulators, which may introduce i.i.d. bias and limit generalization. Second, our reward design does not explicitly penalize excessive or unnecessary clarification questions. In different settings, this could encourage reward hacking, leading the agent to ask overly long or repetitive questions, potentially reducing user satisfaction in real-world deployments~\citep{levandovsky2025learning}. Addressing this trade-off between clarification utility and user burden is an important direction for future work.

\paragraph{Future Work}
Looking ahead, future directions include developing multi-task reward functions that jointly optimize for clarification, task execution, and response quality by extending RLVR beyond clarification to broader collaborative reasoning. Another promising direction is teaching tool-use~\citep{qian2025toolrl} and clarification skills with RLVR in multi-turn conversations~\citep{acikgoz-etal-2025-single} in dynamic environments. 
Finally, self-improving LLM agents represent a promising and largely underexplored direction~\citep{Schmidhuber2007}, especially for TOD Agents. 
Future work can focus on enabling agents to proactively self-improve their skills at test time~\citep{acikgoz2025ttsi}, allowing them to adapt to new situations and better align with human preferences on the fly~\citep{carroll2024ai}. 
Beyond purely autonomous agents, an even safer and more compelling direction is the co-evolution of agents together with humans, where continual mutual adaptation enables more reliable, aligned, and effective AI systems~\citep{weston2025ai}.
Together, these efforts move toward a unified objective: building interactive conversational agents capable of reasoning, clarifying, and acting toward perfect collaboration.

\bibliography{custom}

\clearpage
\appendix
\section*{Appendix}

\section{Collaborative Environment}
\label{app:env}
\noindent\textbf{Overview.} To simulate realistic user environments, agents must engage in collaborative communication that handles real-world goal-oriented tasks, where a single user request may encompass several tasks from different domains with varying levels of complexity. In SpeakRL, agents communicate with users in a realistic end-to-end manner, where an agent can directly respond with natural language, take actions via APIs by interacting with external databases, or ask clarification questions.

\vspace{1mm}

\noindent\textbf{Task Generation.} We generate tasks using the user goals $G$ from MultiWOZ 2.4~\citep{ye2022multiwoz24}, which provides ground truth user goals as annotations. Our environment includes five different domains: restaurant, hotel, train, attraction, and taxi. The agent must track user multi-intent goals, monitor the evolving belief state, make API calls when necessary, ask clarification questions in cases of ambiguity or underspecification to advance the task, and provide appropriate system responses (see \Cref{tab:dataset} for further details about environment).

\vspace{1mm}

\noindent\textbf{Collaborative Conversation.} We simulate conversations between an agent and a human user-simulator~\citep{xu2024autotod}, which has access to user goals unknown to the agent. The agent's task is to fulfill the user request, which may involve several different subtasks (e.g., booking a hotel, searching for an Italian restaurant afterward, and reserving a table for 3 persons at 7pm), some of which may include ambiguities requiring user clarification. The agent should interact with the user, gather all necessary information, and complete the task. The task is considered complete when the agent returns the correct booking or reservation IDs, or terminated after specific number of turns.

\section{Further Details on MultiWOZ 2.4 } 
\label{app:evaluation}
\vspace{-1mm}

We evaluate the performance of our SpeakRL using dialogue-level metrics that capture both the effectiveness and efficiency of task completion. Our primary metric is Success Rate, which measures whether the agent fully satisfies all user-specified constraints and successfully completes the task. For each dialogue, we use an LLM-based judge to assess if the agent’s final response fulfills every requirement defined by the user’s goal, including both requested attributes (such as hotel name or train arrival time) and booking constraints (such as the number of people or destination) following \citet{xu2024autotod}. Formally, a dialogue is considered successful if all constraints in the user’s goal $G$ are met by the end of the interaction: $\text{Success} = \mathbb{I}(\text{all constraints in } G \text{ are satisfied})$, where $\mathbb{I}(\cdot)$ denotes the indicator function. This score is computed for every dialogue and averaged across the evaluation set. To account for the stochastic nature of both model inference and LLM-based judging, we conduct five independent runs for each experimental configuration. 

We report two aggregate Success Rate metrics: \textbf{Success Avg@5}, the mean and standard deviation of success rates over the five runs, providing a robust measure of typical performance and variance, and \textbf{Average Number of Turns} per conversation as an efficiency metric. This measures the average length of the dialogue required to complete the task, with lower values indicating more concise and effective interactions.

\section{RLVR Training Details } 
\label{app:rlvr-training}
We conduct our experiments using the TRL framework\footnote{\url{https://github.com/huggingface/trl}} with the GRPO class. We adopt the training prompt template shown in \Cref{tab:deepseek_template} and report the GRPO hyperparameter settings in \Cref{tab:grpo-hyperparams} to ensure reproducibility. The LaaJ prompt template used by the reward model (illustrated in \Cref{fig:grpo}) is provided in \Cref{fig:laaj}.

\begin{table}[!h]
\centering
\small
\setlength{\tabcolsep}{6pt}
\begin{tabular}{l l}
\toprule
\textbf{Hyperparameter} & \textbf{Value} \\
\midrule
Base Model & Qwen/Qwen2.5-7B-Instruct \\
Dataset & SpeakER 25K \\
\midrule
Epochs & 1 \\
Batch Size (per device) & 8 \\
Gradient Accumulation Steps & 8 \\
Effective Batch Size & 512 \\
Learning Rate & $1 \times 10^{-5}$ \\
LR Scheduler & Cosine \\
Warmup Ratio & 0.1 \\
Optimizer & AdamW \\
Adam $\beta_1$ and $\beta_2$ & 0.9, 0.99 \\
Weight Decay & 0.1 \\
Max Gradient Norm & 0.1 \\
\midrule
GRPO $\beta$ & 0.04 \\
Number of Generations ($K$) & 8 \\
Max Prompt Length & 512 \\
Max Completion Length & 786 \\
Precision & BF16 \\
GPUs & 8 $\times$ A100s \\
\bottomrule
\end{tabular}
\caption{GRPO training hyperparameter details.}
\label{tab:grpo-hyperparams}
\end{table}

\begin{table*}[h!]
\centering
\resizebox{\textwidth}{!}{%
\begin{tabular}{lrcc}
\toprule
\textbf{Domain} & \textbf{API Name}   & \textbf{API Arguments}                                          & \textbf{Test Samples per Domain}   \\ \midrule
Restaurant      & query\_restaurant   & area, pricerange, food, name                                    & 437   \\               
                & book\_restaurant    & name, people, day, time, pricerange, stars, type                &    \\ \midrule
Hotel           & query\_hotel        & area, internet, name, parking                                   & 394        \\
                & book\_hotel         & name, people, day, stay                                         &          \\ \midrule
Attraction      & query\_attraction   & area, name, type                                                & 395   \\ \midrule
Train           & query\_train        & arriveBy, day, departure, destination, leaveAt, trainID         & 494       \\
                & buy\_train\_ticket  & arriveBy, day, departure, destination, leaveAt, trainID, people &   \\ \midrule
Taxi            & book\_taxi          & arriveBy, departure, destination, leaveAt                       & 195          \\ \bottomrule
\end{tabular}%
}
\caption{Environment details and available function calls.}
\label{tab:dataset}
\end{table*}

\begin{figure*}[t]
\centering
\begin{tcolorbox}[title=\textbf{LLM Judge Prompt for Quality Reward}, label={prompt:llm_judge_prompt}]
\small
You are a judge evaluating the quality of user clarification questions. Given a conversation agent clarification question, analyze if there are any clarification questions and evaluate their quality.

\textbf{Rules:}
\begin{enumerate}[leftmargin=*, topsep=2pt, itemsep=2pt]
    \item If clarification questions exist, evaluate them based on:
    \begin{itemize}[leftmargin=1.2em, itemsep=1pt]
        \item Relevance to the context
        \item Precision and clarity
        \item Specificity
        \item Logical connection to previous context
        \item Constructive nature of the question
    \end{itemize}
    \item If no clarification questions exist, output: \texttt{0}
    \item Output format:
    \begin{itemize}[leftmargin=1.2em, itemsep=1pt]
        \item For high-quality clarification questions: \texttt{1}
        \item For low-quality or no clarification questions: \texttt{0}
    \end{itemize}
\end{enumerate}

\textbf{IMPORTANT:} You must \textit{only} output the number \texttt{0} or \texttt{1}. No other text, explanations, or characters are allowed. Do not provide any reasoning. Return only an integer score in the following exact format:
\\ \\
Score: [YOUR BINARY 0/1 SCORE HERE]
\\ \\
\textbf{Conversation}\\
\texttt{<}conversation\texttt{>}
\\ \\
\textbf{Agent Clarification Question to Judge}\\
\texttt{<}clarification\_question\texttt{>}
\\ \\
\textbf{Your Decision (0/1)} \\
\textbf{Score:} [0 or 1]

\normalsize
\end{tcolorbox}
\caption{LLM Judge prompt used for binary quality reward evaluation.}
\label{fig:laaj}
\end{figure*}

\end{document}